\documentclass[10pt,twocolumn,a4paper]{article}
\usepackage{times} 
\usepackage{helvet} 
\usepackage{courier} 
\usepackage{url}
\usepackage{latexsym}
\usepackage{verbatim}
\usepackage{multirow}
\usepackage{float}
\restylefloat{table}
\usepackage{graphicx}
\usepackage{algpseudocode}
\usepackage{algorithm}
\usepackage{color}
\usepackage{amscd}
\usepackage{amsmath}
\usepackage{amssymb}
\usepackage{tabularx}
\usepackage{relsize}
\usepackage{mdframed}
\usepackage{bm}
\usepackage{enumitem}
\usepackage{caption}
\usepackage{subcaption}
\newcommand\Mark[1]{\textsuperscript#1}
 
\begin{document}

\title{Inferring Interpersonal Relations in Narrative Summaries}
\date{}

\author{
  Shashank Srivastava\Mark{1}, Snigdha Chaturvedi\Mark{2}, Tom Mitchell\Mark{1}\\
  \Mark{1}Carnegie Mellon University \Mark{2}University of Maryland, College Park\\
  ssrivastava@cmu.edu, snigdhac@cs.umd.edu, tom.mitchell@cmu.edu
}

%\end{comment}

\maketitle

\begin{abstract}
Characterizing relationships between people is fundamental for the understanding of narratives. In this work, we address the problem of inferring the polarity of relationships between people in narrative summaries. We formulate the problem as a joint structured prediction for each narrative, and present a model that combines evidence from linguistic and semantic features, as well as features based on the structure of the social community in the text. We also provide a clustering-based approach that can exploit regularities in narrative types. e.g., learn an affinity for love-triangles in romantic stories. On a dataset of movie summaries from Wikipedia, our structured models provide more than a 30\% error-reduction over a competitive baseline that considers pairs of characters in isolation.
\end{abstract}
\section{Introduction}
\label{introduction}

\setlength{\textfloatsep}{20pt}
Understanding narratives requires the ability to interpret character intentions, desires and relationships. The importance of characters and characterization in narratives has been explored in recent works that focus on their roles and representations \cite{bamman2014bayesian,valls2014toward,chambers2013event}, as against a plot-centric perspective of a narrative as primarily a sequence of events \cite{finlayson2012learning,schank1977scripts,chambers2008unsupervised}.
However, while such approaches can identify characters types, they do not model \emph{relationships} between characters in a narrative.

In this work, we address the problem of inferring cooperative and adversarial relationships between people in narrative summaries. Identifying character cooperation and conflict is essential for narrative comprehension. It can guide interpretation of narrative events, explain character actions and behavior and steer the reader's expectation about the plot. As such, it can have value for applications such as machine reading, QA and document summarization.

\begin{figure}[h]
    \centering
    \footnotesize{
        \begin{mdframed}
        \noindent \emph{ Young drifter Axel Nordmann %arrives at the waterfront on the west side of Manhattan, seeking employment as a longshoreman. He 
        goes to work in a gang of stevedores headed by Charlie Malik, a vicious bully, and is befriended by Tommy Tyler, who also supervises a stevedore gang. Malik resents blacks in positions of authority, and is antagonized when Axel goes to work for Tommy. Axel moves into Tommy's neighborhood and becomes friends with Tommy's wife Lucy. % and becomes romantically involved with her friend Ellen. 
        Axel is hiding something, and it emerges that he is a deserter from the United States Army. 
        Malik is aware of that, and is extorting money from him. Malik frequently tries to provoke Tommy and Axel into fights, with Tommy coming to Axel's aid ...}
        \end{mdframed}

        \subcaption{}
        \label{fig:1a}
    }
    {
        \includegraphics[scale=0.25]{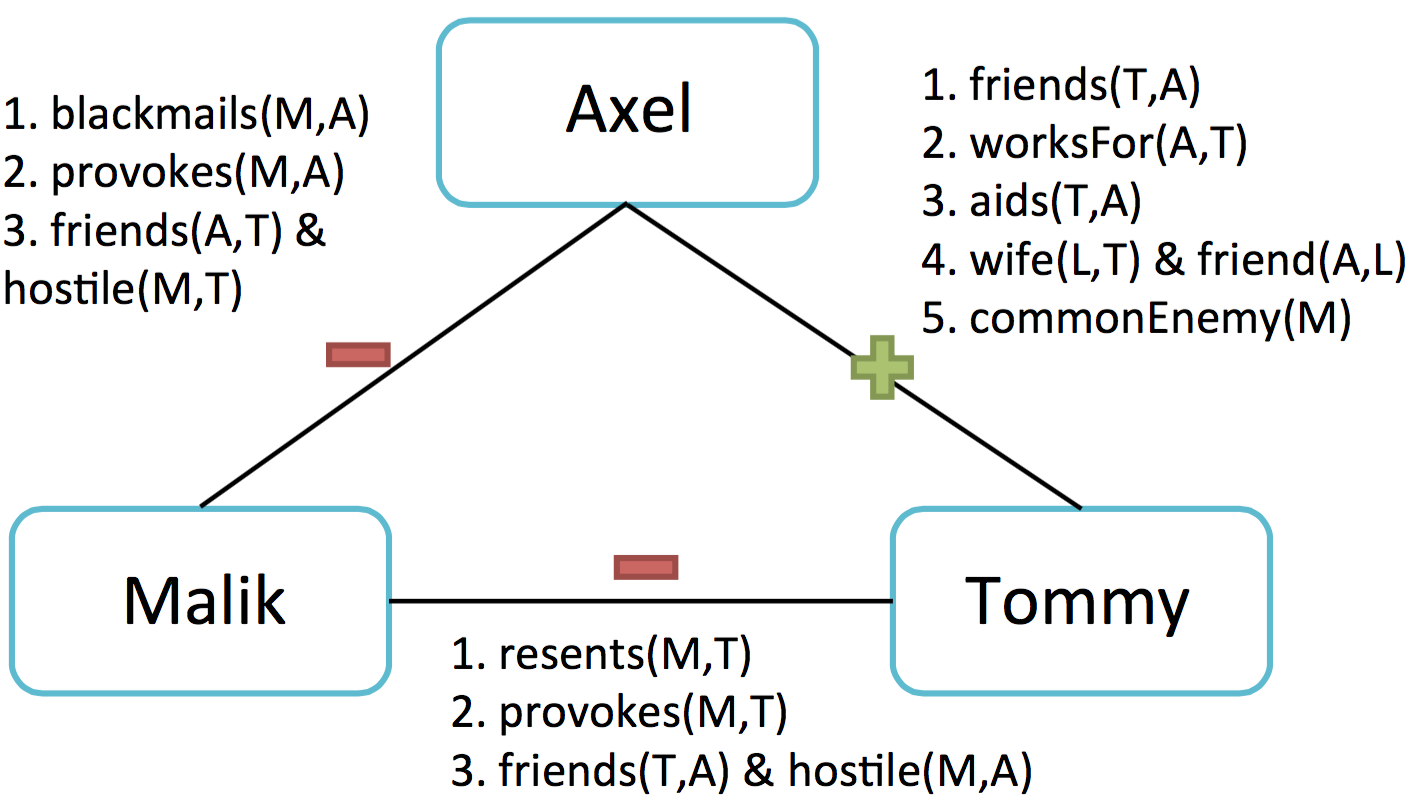}
        \subcaption{}
        \label{fig:1b}
    }
  \caption{ (a) Sample summary extract for the 1957 movie `Edge of the City'; and (b) inferred relationship polarities with supporting evidence}
\end{figure}

As a motivating example, let us consider the plot summary in Figure %\ref{summary_edge}
\ref{fig:1a} (condensed here for brevity). In this passage, the relations between the principal characters are explicated through a combination of %linguistic, syntactic and semantic 
cues, as seen in Figure \ref{fig:1b}. For instance, one can infer that Alex (A) and Tommy (T) have a cooperative relationship through a combination of the following observations (among others): (1) T initially `befriends' A, (2) A works for T, and its connotation that A is likely to cooperate with T , (3) T aids A in fights, (4) A is a friend of T's wife %\footnote{ This might be detrimental to A and T's mutual relation, if this relation between T and A's wife was romantic.}
, (5) A and T have a common adversary. %The ability to make such an inference can assist understanding of the text, as well as predict/justify future actions of A or T in the text (for example, this might explain why A later kills M in the story). 
In particular, we note that cues (4) and (5) cannot be extracted from looking at the relation between A and T in isolation, but depend on their relations with others. In this work, we show that % apart from linguistic evidence, 
such indirect structural cues can be very significant for inference of character relationships. 
%Even this simple example displays some difficulties of the problem. The complexities of narrative, discourse and issues of entity-tracking and coreference suggest the need for a rich enough set of features that can faithfully indicate social dynamics in the text.

Our problem formulation assumes a fixed relation between a pair of characters within a narrative. While this can be problematic since relationships can transform over time; in a wide range of examples, the assumption is reasonable. Even in complex narratives, relationships remain persistent within sub-parts. From a pragmatic perspective, the approximation serves as a useful starting point for research. Our main contributions are:

\begin{itemize}[leftmargin=*]
\item We introduce the problem of characterizing cooperation in interpersonal relations in narrative texts; and provide an annotated dataset of 153 movie summaries for the task. %with relationships labeled on a predefined ordinal scale.
\item We design linguistic, semantic and discourse features for this task, and demonstrate their relative importance. %for the task treating the problem as a classification task for each pair of characters
\item We formulate the problem as structured prediction, and introduce a joint model that incorporates both text-based cues and structural inferences.
%\item We provide a clustering-based extension to exploit regularities in narratives, enabling content-specific models. \footnote{e.g., predict more hostile relations in revenge-dramas, rather than love stories} % based on observed regularities in the data. 
\item We provide an extension to exploit narrative-specific regularities, enabling content-based models. \footnote{e.g., predict more hostile relations in revenge-dramas, rather than love stories} % based on observed regularities in the data.
\end{itemize}
%Our models and features lead to accurate relation predictions, as well as meaningful and interpretable justifications in our experiments.
The layout of the paper is as follows: In section \ref{method}, we present our models, and formulation of the problem as a structured prediction. In section \ref{features}, we describe the %text-based 
features used by our models in detail. We then describe our dataset, and present quantitative and qualitative evaluations of our models. %Finally, we place our model in perspective of related work; and conclude with a discussion of limitations of our approach as well as promising directions of future work.

\setlength{\textfloatsep}{14pt}
\section{Related work}

Existing research on characterizing relationships between people has almost exclusively focused on dialogue or social network data.
%\cite{bramsen2011extracting,hassan2012extracting,krishnan:2015}
Such methods have explored aspects of relations such as power \cite{bramsen2011extracting}, address formality \cite{krishnan:2015} and sentiment \cite{hassan2012extracting} in conversations. Recently, \cite{agarwal2014parsing} studied the problem of parsing movie screenplays for extracting social networks.
However, analysis of character relationships in narrative texts has largely been limited to simplistic schemes based on counting character co-occurrences in quoted conversations \cite{elson2010extracting} or social events \cite{agarwal2013automatic}. 
%Studies of character relationships in narrative text have focused on building social networks between literary characters based on frequencies of character conversations occurrences \cite{agarwal2013automatic}, co-occurrence in narrative events and attributes such as centrality for literary fiction \cite{elson2010extracting}. Recently, \cite{agarwal2014parsing} studied the problem of parsing movie screenplays for extracting social networks. Among approaches identifying polarity in relations based on linguistic content, \cite{bramsen2011extracting} use N-gram features to predict power-relationships from email data, \cite{hassan2012extracting} predict online forum interactions as positive or negative.
We believe this is the first attempt to infer relation polarities in narrative texts.

In terms of approach, our use of structural triads as features is most closely related to \cite{krishnan:2015} who use an unsupervised joint probabilistic model of text and structure for the task of inducing formality from address terms in dialogue, and \cite{leskovec2010signed} who empirically analyze signed triads in social networks from a perspective of structural theories. Such social triads have previously been studied from perspectives of social psychology and networks \cite{heider1946attitudes,cartwright1956structural}.

\section{Relation classification as Structured Prediction}

\label{method}

%\subsection{Relation classification as Structured Prediction}
%\label{structuredmodel}
We formulate the problem of relation classification to allow arbitrary text-based and structural features. We consider the problem as a structured prediction, where we \emph{jointly} infer the collective assignment of relations-labels for all pairs of characters in a document. Let $\bf x$ denote a narrative document for which we want to infer relationship structure $\bm{y}$. We could think of $\bf x $ as a graph with characters as nodes, and relationship predictions corresponding to edge-labels. We assume a supervised learning setting where we have labeled training set $\mathcal{T}:=\{(\bf{x_1},\bf{y_1}),...,(\bf{x_m},\bf{y_m})\}$. For each $\bm{x}$, we have a set of allowed assignments $\mathcal{Y}(\bm{x})$ (consisting of combinations of binary assignments to each edge-label in $\bm{x}$). % a narrative text $\bm{x}$ and a label-assignment $\bm{y}$, w
Following standard approaches in structured classification, we consider linear classifiers of form:

\begin{equation}
\label{decode}
h_{\bm{w}} = \underset{\bm{y} \in \mathcal{Y}(\bm{x}) }{\operatorname{argmax}} \quad \bm{w}^T \bm{\phi}(\bm{x},\bm{y})
\end{equation}

\noindent Here, $\bm{\phi}(\bm{x},\bm{y})$ is a feature vector that can depend on both the narrative document $\bf x$ and a relation-polarity assignment $\bf y$, $\bm{w}$ is a weight vector, and $\bm{w}^T \bm{\phi}(\bm{x},\bm{y})$ denotes a linear score indicating the goodness of the assignment. %Finding the best assignment then 
%For a given weight vector $\bm{w}$, f
Finding the best assignment corresponds to the decoding problem, i.e. finding the highest scoring assignment under a given model. %, specified by $\bm{w}$.
On the other hand, the model parameters $\bm{w}$ can be learnt using a voted structured perceptron training algorithm \cite{collins2002discriminative}. The structured perceptron updates can also be seen as stochastic sub-gradient descent steps minimizing the following structured hinge loss: \\ %This can be seen as minimizing a structured hinge loss: 
\begin{equation}
\label{perceptronloss}
L(\bm{w},\bm{x},\bm{y}) := \underset{\bm{y'} \in \mathcal{Y}(\bm{x})} {\textrm{max}}  \bm{w}^{T} \big( \bm{\phi}(\bm{x},\bm{y'}) - \bm{\phi}(\bm{x},\bm{y}) \big)
\end{equation}

\noindent For our problem, we define the feature vector $\bm{\phi}(\bm{x},\bm{y})$ as a concatenation of features based on text and structural components: $\bm{\phi}(\bm{x},\bm{y}) := 
\big( 
    \bm{\phi}_{text}(\bm{x},\bm{y}) \;
    \bm{\phi}_{struct}(\bm{x},\bm{y})
\big)$. The text-based component can be defined by extending the traditional perceptron framework %\cite{rosenblatt1958perceptron}
as $\bm{\phi}_{text}(\bm{x},\bm{y}) := \underset{x_e \in E(\bm{x})}{\sum} y_e \bm{\phi}(x_e)$. Here $E(\bm{x})$ consists of the set of annotated character-pair relationships for the narrative text $\bm{x}$, $\bm{\phi}(x_e)$ denotes the text-based feature-representation for the character-pair (as described in Section \ref{textmodel}), and $y_e$ is the binary assignment label ($\pm1$) for the pair in $\bm{y}$.
On the other hand, our structural features $\bm{\phi}_{struct}(\bm{x},\bm{y})$ focus on configurations of relationship assignments of triads of characters, and are motivated in our discussion of transitive relations %and structural balance 
in Section \ref{transitive} . %Social triads have been studied from perspectives of social psychology and network theory\cite{heider1946attitudes,cartwright1956structural}; and we briefly characterize the primary triadic features with our informal appellations for them in Figure \ref{triads}. Values of these features consist of the number of such configurations in any assignment. The affinities for such configurations, as reflected in corresponding learnt weights $\bm{w}_{struct}$ can then be learnt from the data. 
%While the structural features used in our current work only depend on assignment labels of triads of characters, 
We note that while this is not the case in the current work, structural features can also encode character attributes (such as age or gender) in conjunction with assignment labels $\bm{y}$.

\noindent \textbf{Learning and Inference:} Structured perceptrons have been conventionally used for simple structured inputs (sequences and trees) and locally factoring models, which are amenable to efficient dynamic programming inference algorithms. This is because updates require inference over an exponentially large space (solving the decoding problem in Equation \ref{decode}), and updates from inexact search can violate convergence properties. However, \cite{huang2012structured} show that exact search is not needed for convergence as long as we can guarantee updates on `violations' only, i.e. %when the 1-best linear score from the model is higher than the model score for the correct assignment. Hence 
it suffices to find a labeling assignment with higher score than the correct update. Additionally, edge labels are expected to be relatively sparse for our domain since character graphs in most narratives are not fully connected. Hence, the inference problem decomposes for relation-edges which are not parts of structural triangles% in provided annotations
, and the decoding problem can be exactly solved for the vast majority of narrative texts.

%\begin{comment}
\begin{algorithm}[]
  \caption{Perceptron Training for Relations}
  \begin{algorithmic}[1]
%    \State \textbf{Input:} Partially labeled narrative texts $\mathcal{T}= \{ (\bm{x},\bm{y})_{m} \}$
    \State Initialize $\bm{w}$ to $\bm{0}$%, $\forall i$
    \For {$iter:1$ to $numEpochs$}
    \For {$j:1$ to $m$}
        \State Randomly choose $i \in \{1..m\}$\\
        $\quad \quad \quad \bm{\hat{y}} \leftarrow Decode(\bm{x}_i, \bm{w})$
        \If { ( $w^{T} \bm{\phi}(\bm{x}_i,\bm{\hat{y}}) \geq 
             w^{T} \bm{\phi}(\bm{x}_i,\bm{y}_i) ) $}\\
            $\quad \quad \bm{w} \leftarrow \bm{w} + \eta (\phi( {\bf x}_{i}, {\bf y}_{i}) - \phi( {\bf x}_{i}, {\bf y'}) )$
%        \Else
%            No Update
        \EndIf
    \EndFor
    \EndFor
%    \State return $\bm{w}$
  \end{algorithmic}
\end{algorithm}
%\end{comment}

For inference on a new document where the edge relations are not known, decoding can proceed by initializing the narrative graph to high confidence edges from the text-based model only (character relationships firmly embedded in text), and appending single edges which complete triads. To avoid speculative inference of relations between character pairs that are  ungrounded in the text, we only consider structural triads for which at least two edges are grounded in the text while decoding with the structural model.

%\subsection{Content-specific clustering model }
 \subsection{Accounting for narrative types }
 \label{narrativetypes}
 The framework described in the previous section provides a simple model to incorporate text-based and structural features for relation classification. However, a shortcoming in the approach is that the model is agnostic to narrative types. Ideally, a model could allow differential weights to features depending on the narrative type. As speculative illustrations, `Mexican standoffs' might be common in `revenge/gangster' narratives, or family-relations might be highly indicative of cooperation in children stories; and a model would ideally %ideally learn such regularities on its own.
 learn and leverage such regularities in the data.
 
We present a clustering-based extension to our structured model, which can incorporate features descriptive of the narrative text to infer regularities, and make content-based predictions. Let us surmise that the data consists of $K$ natural clusters of narrative-types, with a specific structured model for each cluster (specified by weights $\bm{w}_k$). For each narrative text $x$, we associate a vector $f(x)$ that represents content and determines narrative type. Examples of such representations could be keywords for a document, genre information for a movie or novel, topic proportions of words in the text from a topic-model, etc. We model the membership of narrative $\bm{x}$ to the cluster $c_k$ by a softmax logistic multinomial.
\begin{equation}
P(c=c_k ; \bm{x}) = \frac{\exp(\lambda_k^T f(\bm{x}) )}{ \sum_{k'} \exp(\lambda_{k'}^T f(\bm{x})) }
\end{equation}
From our observation of the loss objective for the structured  perceptron in Equation \ref{perceptronloss}, we can define the \emph{expected} loss for a narrative text ($\bm{x},\bm{y}$) under the clustering model as:
\begin{equation}
\mathcal{L}(\bm{x},\bm{y}) = \sum_k{\frac{\exp(\lambda_k^T f(\bm{x})) }{ \sum_{k'} \exp(\lambda_{k'}^T f(\bm{x}))}} \quad L(\bm{w}_k,\bm{x},\bm{y})
\end{equation}
Then the overall objective loss over the training set $\mathcal{T}$ is:
\begin{align}
    \begin{split}
 J &= \sum_i \mathcal{L}(\bm{x}_i,\bm{y}_i) \\
   &= \sum_i \sum_k {\frac{\exp(\lambda_k^T f(\bm{x}_i))  L(\bm{w}_k, \bm{x}_i,\bm{y}_i) }{ \sum_{k'} \exp(\lambda_{k'}^T f(\bm{x}_i))}} 
   %&= \sum_i \sum_k {\frac{\exp(\lambda_k^T \bm{x}_i) \Big( \underset{\bm{y'} \in \mathcal{Y}(\bm{x})} {\textrm{max}}  \bm{w}^{T} \big( \bm{\phi}(\bm{x},\bm{y'}) - \bm{\phi}(\bm{x},\bm{y}) \big) \Big) }{ \sum_{k'} \exp(\lambda_{k'}^T \bm{x}_i)}} 
   \end{split}
\end{align}
We jointly minimize the overall objective through a block-coordinate descent procedure. This consists of a two-step alternating minimization of the objective w.r.t. the prediction model weights $\bm{w}_k$ and the clustering parameters $\lambda_k$, respectively. In the first step, we optimize the prediction model weights $\bm{w}_k$ while fixing the clustering parameters $\lambda_k$. This can be done by weighting the training examples for each cluster by their cluster membership; and invoking the structured perceptron procedure for each cluster. In the alternating step, we fix the predictions model weights; and update the clustering parameters using gradient descent:\\ \\
\footnotesize
$\nabla_{\lambda_k} J = \underset{i=1}{\sum}  \frac{\exp(\lambda_k^T f(\bm{x}_i))}{ Z_i^2 } 
\underset{k'}{ \sum} \exp(\lambda_k'^T f(\bm{x}_i)) (L(\bm{w}_k,\bm{x}_i,\bm{y}_i) - L(\bm{w}_k',\bm{x}_i,\bm{y}_i)) f(\bm{x}_i)$  \\
\normalsize

\begin{comment}
\scriptsize
\begin{equation*}
\nabla_{\lambda_k} J = \underset{i=1}{\sum}  \frac{\exp(\lambda_k^T f(\bm{x}_i))}{ Z_i^2 } 
\underset{k'}{ \sum} \exp(\lambda_k'^T f(\bm{x}_i)) (L(\bm{w}_k,\bm{x}_i,\bm{y}_i) - L(\bm{w}_k',\bm{x}_i,\bm{y}_i)) f(\bm{x}_i)
\end{equation*}
\normalsize
\end{comment}

\begin{algorithm}[]
  \caption{Narrative-specific Model}
  \label{proc2}
  \begin{algorithmic}[1]
    \State Initialize $\lambda_k$ to random vectors
    \State \textbf{repeat}
    \State \textbf{Update perceptron weights:} Train structured perceptron models for each cluster $c_k$, weighting training instance $(\bm{x}_i,\bm{y}_i)$ in $\mathcal{T}$ by $\frac{\exp(\lambda_k^T f(\bm{x}_i))}{ \sum_{k'} \exp(\lambda_{k'}^T f(\bm{x}_i)) }$
    \State \textbf{Update clustering model}
    \For {$i:1$ to $numIter$}
        \For {$k:1$ to $K$}
            $\quad \lambda_k \leftarrow \lambda_k - \mu \nabla_{\lambda_k} J $%\frac{\partial J}{\partial \lambda_k}$
        \EndFor 
    \EndFor
%    \State \textbf{till convergence}
  \end{algorithmic}
\end{algorithm}
This can be interpreted as a bootstrapping procedure, where given cluster assignments of points, we update the prediction model weights; and given losses from the prediction model, update data clusters parameters to reassign the most violating data-points. We note that the objective is non-convex due to the softmax, and hence different initializations of the procedure can lead to different solutions. However, since each sub-procedure decreases the objective value; the overall objective decreases for small enough step sizes. The procedure is summarized in Algorithm \ref{proc2}. 
For prediction, each narrative text is assigned to the most likely cluster with the clustering model. 

To efficiently use training data, we allow  parameter-sharing across cluster-specific prediction models, drawing from methods in multi-task learning \cite{evgeniou2004regularized}. In particular, we model each $\bm{w}_k$ as composed of a shared base model, and additive cluster-specific weights:
\begin{equation*}
\bm{w}_k = \bm{w}_0 + \bm{v}_k
\end{equation*}

\noindent Implementationally, we can do this by simply augmenting cluster-specific feature representations as follows:
\small
$\phi_k(\bm{x},\bm{y}) = 
\Big( 
(1-\alpha) \phi(\bm{x},\bm{y}) , \underbrace{\bm{0},...,\bm{0}}_\text{k-1}, 
\alpha \phi(\bm{x},\bm{y}) ,
\underbrace{\bm{0},...,\bm{0}}_\text{K-k}
\Big)$
\normalsize
Here $\alpha$ is a hyper-parameter between 0 and 1, which specifies the weighting of the shared and cluster-specific models. $\alpha= 0$ negates clustering, and reduces the clustering model to the plain structured model without clustering. % in Section \ref{structuredmodel}. 
Conversely, $\alpha=1$ implies no parameter sharing across clusters.

\section{Features}

\label{features}

In this section, we outline the text-based and structural features used by our classification models. %
The text-based features make use of existing linguistic and semantic resources, whereas the structural features are based on counts of specific signed social triads, which can be enumerated for any assignment. %such as connotation lexicons, multi-word-expression databases, frame-semantic and syntactic parses; as well as features based on gender and discourse. On the other hand, the structural features are based on social triads.
We provide implementations for these features, as well as the complete pipeline for annotating relationship polarities for a new text document on our project webpage. 

\subsection{Text-based cues ($\bm{\phi_{text}}$)}
\label{textmodel}
These features aim to identfiy relationships between pairs of characters in isolation. %For this task, we extract a set of features to represent different aspects of their relations. Our features 
These are based on resources such as sentiment lexicons, syntactic and semantic parsers, distributional word-representations, and multi-word-expression dictionaries, and % The features 
are engineered to capture:

\begin{itemize}[noitemsep,nolistsep]
\item{Overall polarities of interactions between characters (from text-spans between coreferent mentions) based on lexical and phrasal-level %subjectivity 
polarity clues.}
\item{Semantic connotations of actions one agent does to the other, actions they share as agents or patients, and how often they act as a team.} %they do to each other, actions they share as agents or recipients, and how often they act as a team (from syntactic parses)}
\item{Character co-occurrences in semantic frames that evoke positivity, negativity or social relationship .} %(from frame-semantic parses).}
\item{Character similarity based on whether they are described by similar adjectives; and the narrative sentiment of adverbs describing their actions.}
\item{Existence of familial relations between characters. }
\end{itemize}
%(1) overall polarities of interactions between characters (from text-spans between coreferent mentions) based on lexical and phrasal-level subjectivity clues; (2) semantic connotations of actions they do to each other, actions they share as agents or recipients, and how often they act as a team (from syntactic parses); (3) their co-occurrences in semantic frames indicating positivity, negativity or social relationship (from frame-semantic parses); (4) character similarity based on adjectives describing characters, and sentiment of adverbs describing shared acts; and (5) existence of family relations between them. %; and subjectivity clues based on phrase-level sentiment analysis. 

We base our entity-tracking %(for identifying coreferent mentions) 
on the Stanford Core-NLP system; and augment the computation of all sentiment features with basic negation handling. %\footnote{ We provide implementations for these features, as well as the complete pipeline on our project webpage } 
Based on such features extracted for each character pair, relationship characterization can be treated as supervised classification (with $y=\pm1$ corresponding to cooperative or adversarial relations). Our baseline unstructured approach uses only these features with a logistic regression model.\\
%our unstructured text-based model trains a binary classifier for predicting cooperation in the relationship using common supervised learning algorithms.

\noindent \textbf{Feature details:} Texts are initially processed with the Stanford Core NLP system to identify personal named entities, and obtain dependency parses. As basic post-processing, we mark coreferent text-spans with their corresponding personal entities, using basic string-matching and pronominal resolution from the Stanford coreference system. For enumerating actions by/on two characters of interest, we identify verbs in sentences for which the characters occurred in a agent or a patient role (identified using `nsubj' or `agent'; and  `dobj' or `nsubjpass' dependencies respectively). We extend this for conjoint verbs, identified by the `conj' relation (e.g., `A shot and injured B'). The dependency relation `neg' is used to determine negation status of verbs. 

Given a pair of characters, we identify the set of sentences which contain mentions of both (identified by coreferent text-spans). For this set, we extract the arithmetic means, maximum and cumulative sums for the following %families 
sentence-level cues as text-based features (whenever meaningful): 

\begin{figure}[]
\centering
\includegraphics[scale=0.22]{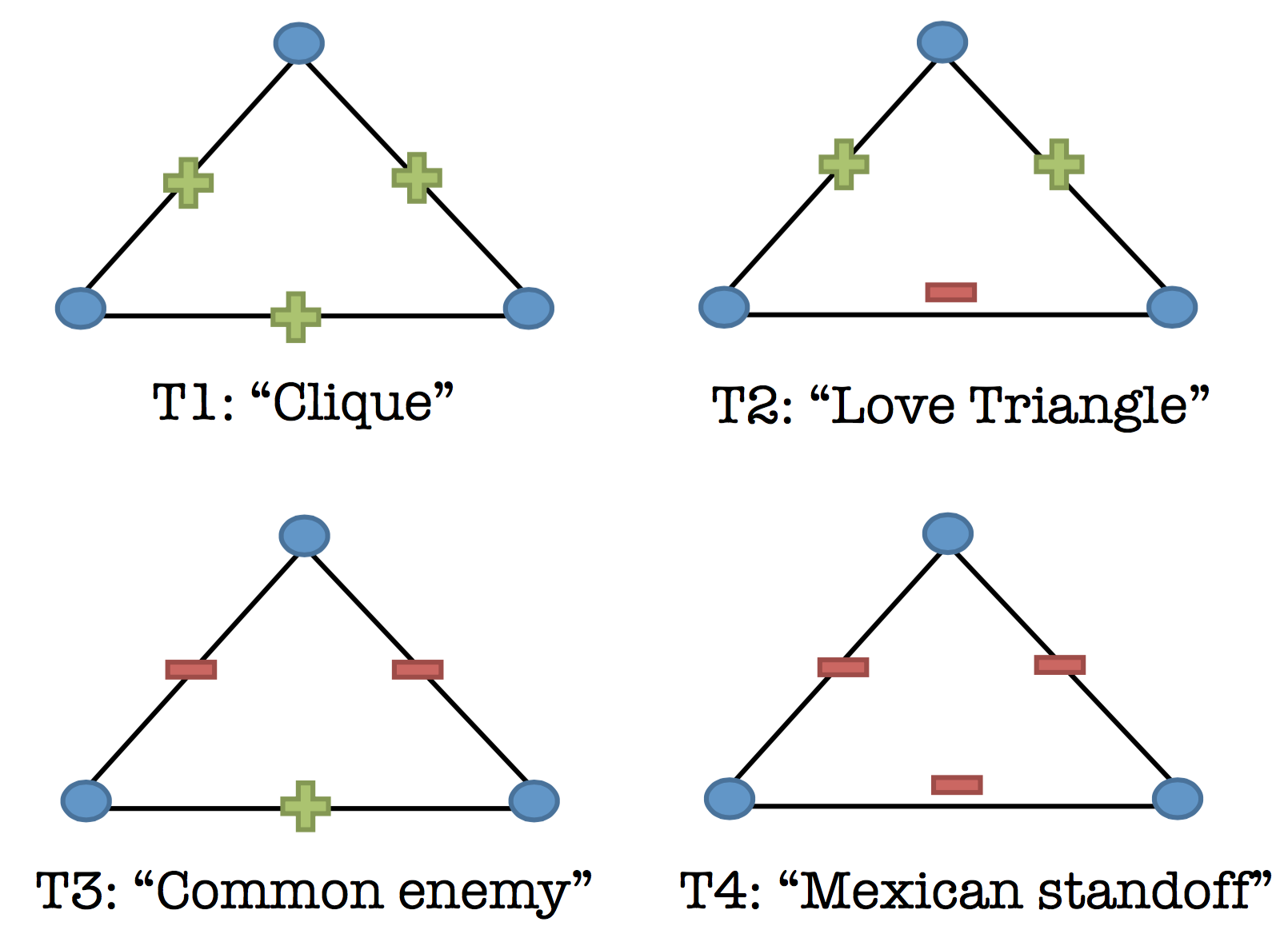}
\caption{Primary triadic structural features. `+' signs indicate cooperative and `--' indicate adversarial relationships}
\label{triads}
\end{figure}

%\noindent \textbf{1. Parse based:} We analyze actions by/on the two characters by identifying all verbs in the sentence, for which the characters occurred in a agent or a patient role (identified using `nsubj' and `agent'; and  `dobj' and `nsubjpass' dependency relations respectively). We extend this for verbs conjoined with each other, identified by a `conj' relation (e.g., A shot and injured B). We used the dependency `neg' relation to determine the negation status of these verbs. Based on this information we extracted the following features:
%\begin{changemargin}{0.2cm}{0cm} 
\begin{enumerate}[leftmargin=*,noitemsep,nolistsep]

\item {\textit{Are Team:} This models if the two characters participate in acts together. It is a binary feature indicating if the two characters were both agents (or patients) of a verb in a sentence e.g., `Malik provokes Tommy and Axel'. }

\item {\textit{Acts Together:} These features count the number of actions with positive and negative connotations that either character (in an agent role) does to the other (in a patient role). There are three variants based on different word connotation resources, viz., semantic lexicons for subjectivity clues~\cite{connotation:2013}, sentiment~\cite{Liu:2005} and prior-polarity~\cite{Wilson:2005} of verbs. The feature does not fire for neutral verbs. e.g, `Malik blackmails Axel'.}

\item {\textit{Surrogate Acts Together:} 
Coverage for the above features suffers from limitations of NLP processing tools. e.g., In `On being provoked by Malik, Tommy...' , Tommy is not a direct patient of the verb. These features extend coverage to verbs which have either of the characters as the agent or the patient in sentences that did not contain any other character apart from the two of interest.}

\item {\textit{Adverb Connotations:} This feature models the narrative's overall bias in describing either character's actions by summing the %the adverbs modifying actions identified above . These features identify the 
semantic connotations of adverbs that modify their joint(or surrogate) acts. e.g., `Tommy nobly befriends Axel'. Positive adverbs count as +1, negative as -1. Uses the same connotation resources as 2. }

\item {\textit{Character similarity:} Models similarity of two characters as the average pairwise similarity of adjectives describing each (where lexical similarity is given by the cosine similarity of embeddings from \cite{collobert2011natural}). This is computed for the entire document, and not at the per-sentence level.}

\item {\textit{Lexical sentiment:} These features count the number of positive and negative sentiments words or multi-word phrases in spans between mentions of the two characters using sentiment lexicons (similar to 2). For multi-word phrases (identified from a list of MWEs), we use a dictionary to map these to single words if possible, and look for these words in connotation lexicons. e.g., `kick the bucket' maps to `die'. This helps with phrases like `fell in love', where `fell' has a negative connotation by itself.}

\item {\textit{Relation keywords:} This feature indicates presence of keywords denoting familial ties between characters (`father', `wife', etc.) in spans between character mentions.}

\item {\textit{Frame semantic:}  These are based on Framenet-style parses of the sentence from the Semafor parser ~\cite{Das:2014}. We compiled lists of frames associated with: (i) positive or (ii) negative connotations (e.g., frames like \emph{cause hurt} or \emph{rescue}), (iii) personal or professional relationships (e.g., %like \emph{subordinates and superiors} or 
\emph{forming relationships}). Three binary features denote frame evocation for each of these lists.} % Our lists also contained ambiguous frames like `cause\_bodily\_experience' in which case the exact connotation of the frame was determined on-the-fly depending on the lexical unit at which that frame fired.}
\end{enumerate}

\subsection{Structural cues ($\bm{\phi_{struct}}$)}
\label{transitive}
As motivated earlier, %in Section \ref{introduction}, 
%aside from clues in the text directly pertaining to relations between characters, 
relationships between people can also be inferred from their relationships with others in a narrative. Our thesis is that a joint inference model that incorporates both structure and text would perform better than one that considers pairwise relations in isolation. 
In some domains, observed relations between entities can directly imply unknown relations among others dues to natural orderings. For example, temporal relations among events yield natural transitive constraints. For the current task; such constraints do not apply. While structural regularities like `a friend of a friend is a friend' might be prevalent, these configurations are not logically entailed; and affinities for such structural regularities must be directly learnt from the observed data. %Rather, However, the incorporation of structural features is not straightforward; since 
%Such social triads have been studied from perspectives of social psychology and network theory\cite{heider1946attitudes,cartwright1956structural}.

In Figure \ref{triads}, we characterize the primary triadic structural features that we use in our models, along with our informal appellations for them. % in Figure \ref{triads}. 
%We incorporate these %primary triads 
%as structural features in our model. 
The values of the four structural features for a narrative document $\mathbf{x}$ and relation polarity assignment $\mathbf{y}$ are simply the number of such configurations in the assignment, and are easily computed. The empirical affinities for such configurations, as reflected in corresponding weights %$\bm{w}_{struct}$ 
can then be learnt from the data.

\begin{comment}
\begin{table}
\footnotesize
\begin{center}
\begin{tabular}{|p{0.22\columnwidth}|p{0.25\columnwidth}|p{0.35\columnwidth}|}\hline
Type & Frame & Frame-elements\\\hline
Negative        & `cause\_harm'                 & `agent', `victim'\\\cline{2-3}
                & `attack'                  & `assailant', `victim'\\\hline
Positive        & `forgiveness'             & `judge', `evaluee'\\\cline{2-3}
                & `supporting'              & `supporter', `supported'\\\hline
Ambiguous       & `cause bodily experience' & `agent', `experiencer'\\\cline{2-3}
                & `friendly or hostile'     & `side\_1', `side\_2', `sides'\\\hline
Relationship    & `kinship'                 & `alter', `ego', `relatives'\\\cline{2-3}
                & `subordinates and superiors' &`superior', `subordinate'\\\hline
\end{tabular}
\caption{Sample of frame categorizations}
\label{table:framesList}
\end{center}
\vspace{-0.1in}
\end{table} 
\end{comment}
\normalsize

%\end{comment}
\section{Dataset}
%We created an annotated dataset of 153 movie summaries from Wikipedia, which were part of the CMU Movie summary corpus\cite{dbamman2013}. Our annotated dataset (consisting of 1044 relation annotations) will be publicly made available for future research.

\begin{figure}[h]
\centering
\includegraphics[scale=0.30]{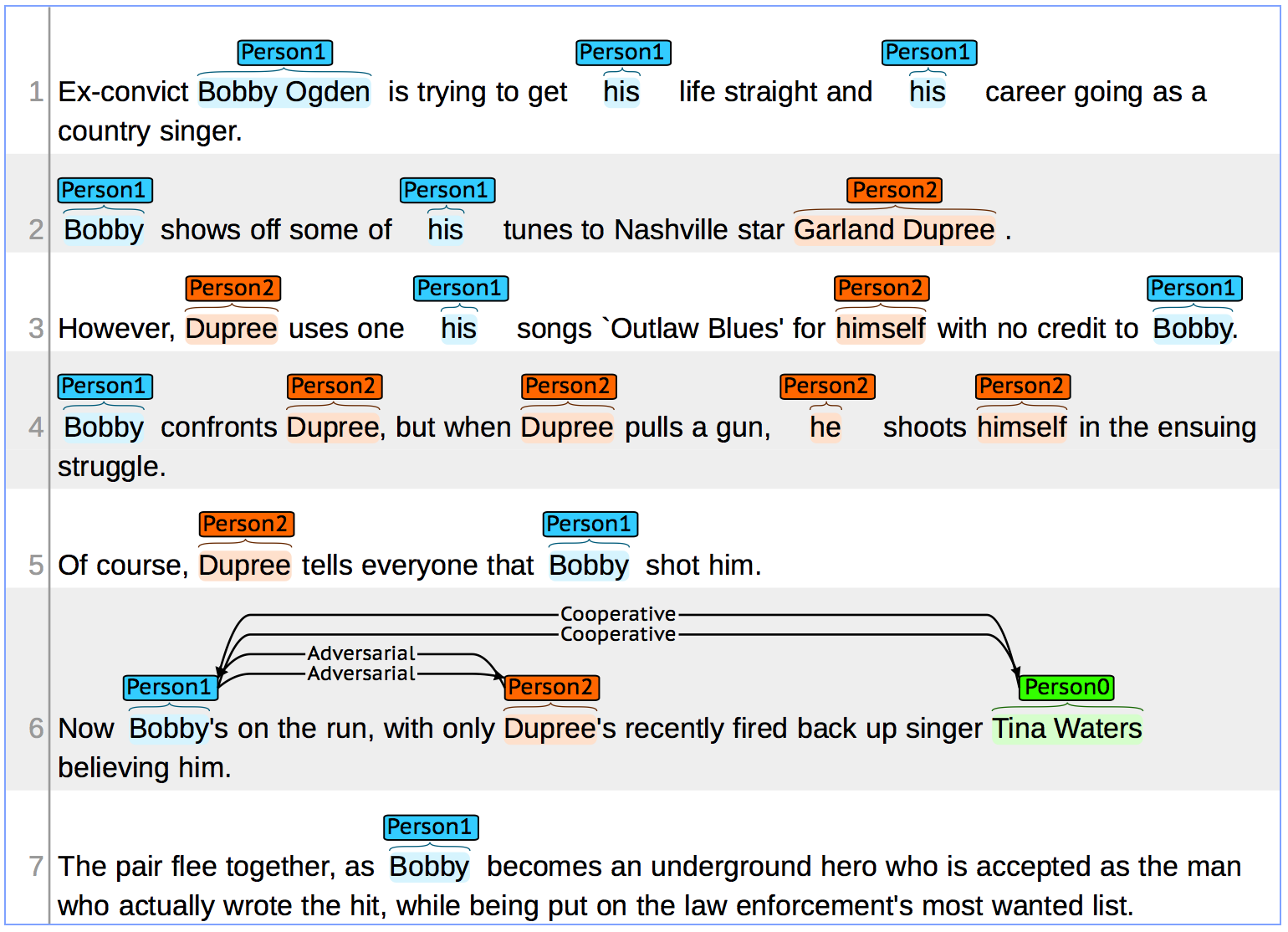}
\caption{Sample annotation for a very short summary}
\label{outlaw_blues}
\end{figure}

%\noindent \textbf{Collection and annotation:} 
We processed the CMU Movie Summary corpus, %which is a collection of 42,306 movies summaries extracted 
a collection of movie plot summaries from Wikipedia, along with aligned meta-data \cite{dbamman2013}; and set up an online annotation task using BRAT \cite{stenetorp2012brat}. We use Stanford Core NLP annotations and basic post-processing to identify personal entities in each text. 

Annotators could choose pairs of characters in the text, and characterize a directed relationship between them on an ordinal five-point scale as `Hostile', `Adversarial', `Neutral', `Cooperative' or `Friendly'. %A sample annotation for a very short summary is shown in Figure \ref{outlaw_blues}. 
This resulted in a dataset of 153 movie summaries, consisting of 1044 character relationship annotations.\footnote{Most relations were annotated symmetrically. For relations with asymmetric labels, we `averaged' the annotations in the two directions to get the annotation for the relation.} The dataset is made publicly available for research on our project webpage.

For %the purpose of 
evaluation, we aggregated `hostile' and `adversarial' edge-labels, and `friendly' and `cooperative' edge-labels to have two classes (neutral annotations were sparse, and ignored in the evaluation). Of these, 58\% of the relations were classified as cooperative or friendly, while 42\% were hostile or adversarial. The estimated annotator agreement for the collapsed classes on a small subset of the data was $>$0.95.
\section{Evaluation and Analysis}
In this section, we discuss quantitative and qualitative evaluation of our methods. %We begin with 
First, we make an ablation study to assess the relative importance of families of text-based features. We then make a comparative evaluation of our methods in recovering gold-standard annotations on a held-out test set of movie summaries. Finally, we qualitatively analyze the performance of the model, and briefly discuss common sources of errors.

\begin{figure}[h]
\centering
\includegraphics[scale=0.32]{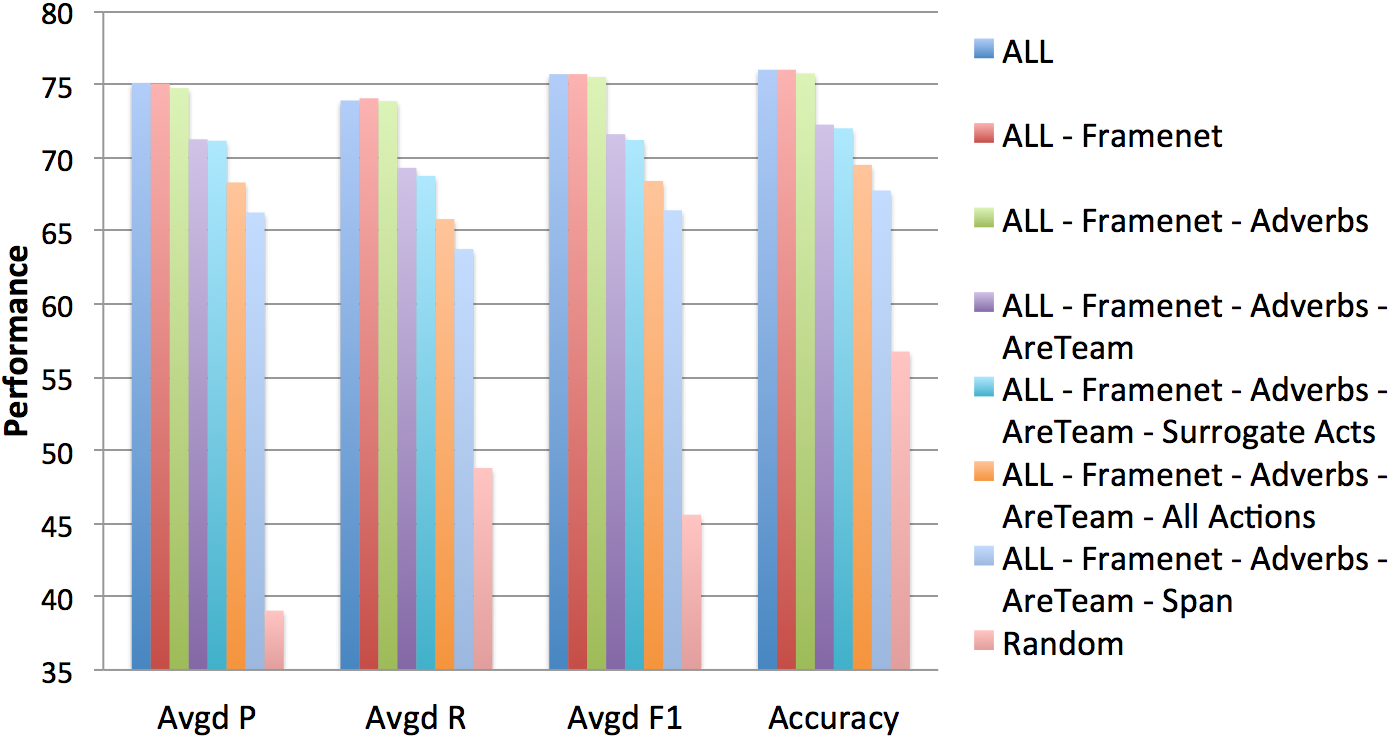}
\caption{Ablation study for text feature families}
\label{ablation}
\end{figure}

\noindent \textbf{Feature ablation:} Figure \ref{ablation} shows the cross-validation performance of major feature families of text features on the training set. We note that Frame-semantic features and adverbial connotations of character actions do not add significantly to model performance. This is perhaps because both these families of features were sparse. Additionally, frame-semantic parses were observed to have frequent errors in frame evocation, and frame element assignment. On the other hand, we observe that joint participation in actions (as agent or patient) is a strong indicator of cooperative relationships. In particular, incorporating these (\emph{`Are Team'}) features was seen to improve both precision and recall for the cooperative class; while not degrading recall for the non-cooperative class. Further, while ignoring sentiment and connotation features for surrogate action features results in marginal degradation in performance; the most significant features are seen to be sentiment and connotation features for actions where characters occur in SVO roles (\emph{`Acts Together'} features); and overall sentiment characterizations for words and phrases in spans of text between character mentions (span based \emph{`Lexical sentiment'} features). %Based on our observations for this study, we do not include Frame-semantic and Adverb based features in our text-based feature-set.

\noindent \textbf{Structured vs unstructured models:}
We now analyze the performance of our proposed models; and evaluate the significance of adding structural features to our models. In our experiments, we found the structured models to consistently outperform text-based models. We tune values of hyper-parameters, i.e. number of training epochs for the structured perceptron (10), the weighting parameter for the clustering model ($\alpha$=0.8), and the number of clusters ($K$=2) through cross validation on training data.\footnote{For representations $f(x)$ of movie summaries, we use genre keywords from the metadata for movies (provided with dataset) and the average of text-feature vectors for all character pairs }
Table \ref{results}  compares the performance of the models on our held-out test set of 27 movie summaries (comprising about 20\% of the all annotated character relations). For the structured models, reported results are averages over 10 initializations. %For this task, we consider the set of edges to be labeled as known. % (this allows the computation of precision and recall for both classes).

We observe that the structured perceptron model for relations (SPR) outperforms the text-only model trained with logistic regression (LR) by a very large margin. These results are consistent with our cross-validation findings, and vindicate our hypothesis that structural features can significantly improve inference of character relations. Further, we observe that the narrative-specific model (with $K=2$) slightly outperforms the structured perceptron model. 

\begin{comment}
\begin{table}[h]
\centering
\begin{tabular}{|l|c|}
\hline
    &   \textbf{Acc}     \\ \hline \hline
LR  &   0.722   \\
SP  &   0.788   \\
CBM &   0.812   \\  \hline
\end{tabular}
\caption{Test set results for relation classification}
\label{results}
\end{table}
\end{comment}

\begin{table}[h]
\centering
\begin{tabular}{l|l|c|l|l}
    &   \textbf{Avg P}   &       \textbf{Avg R}   &  \textbf{Avg F\tiny{1}}  &   \textbf{Acc}     \\ \hline \hline
Naive\footnotesize{ (majority class)} & 0.269  &   0.520   &   0.355    &   0.520   \\
LR  &  0.702    &   0.697   &   0.699    &   0.701   \\
SPR  &  0.794    &   0.793   &   0.793    &   0.792   \\
SPR\footnotesize{ +Narrative types}&  0.806     &   0.804   &   0.805    &   0.804   \\
\end{tabular}
\caption{Test set results for relation classification}
\label{results}
\end{table}

Let us consider the affinities for structural features learnt by the model. Over 10 runs of SPR, the average weights %of the primary structural features %discussed in Figure \ref{triads}
were: $w_{\tt clique}=-2.79$, $w_{\tt love triangle}=-0.84$, $w_{\tt common enemy}=10.26$ and $w_{\tt mexican standoff}= -5.49$. From the perspective of structural balance, the social configurations $\tt love triangle$ and $\tt mexican standoff$ are inherently unstable. %, whereas the others are stable. Hence one might have expected weights to follow this intuition. However, %the learnt weights make a significant departure from this conjecture. Specifically, 
Hence, the learnt affinity for the configuration $\tt love triangle$ seems higher than expected. This is unsurprising, however, if we consider the domain of the data (movies), where it %our informal reference for the configuration this 
might be a common plot element. We also note that the `friend of a friend is a friend' maxim is not supported by the feature weights (even though it is a stable configuration), and hence a model based on this as a hard transitive constraint could be expected to perform poorly.

\begin{comment}
\begin{figure}
\centering
\includegraphics[scale=0.4]{Figures/wtd_triangles}
\caption{Learnt weights of structural features}
\label{wtd_triangles}
\end{figure}
\end{comment}

\begin{figure}
\centering
\includegraphics[scale=0.25]{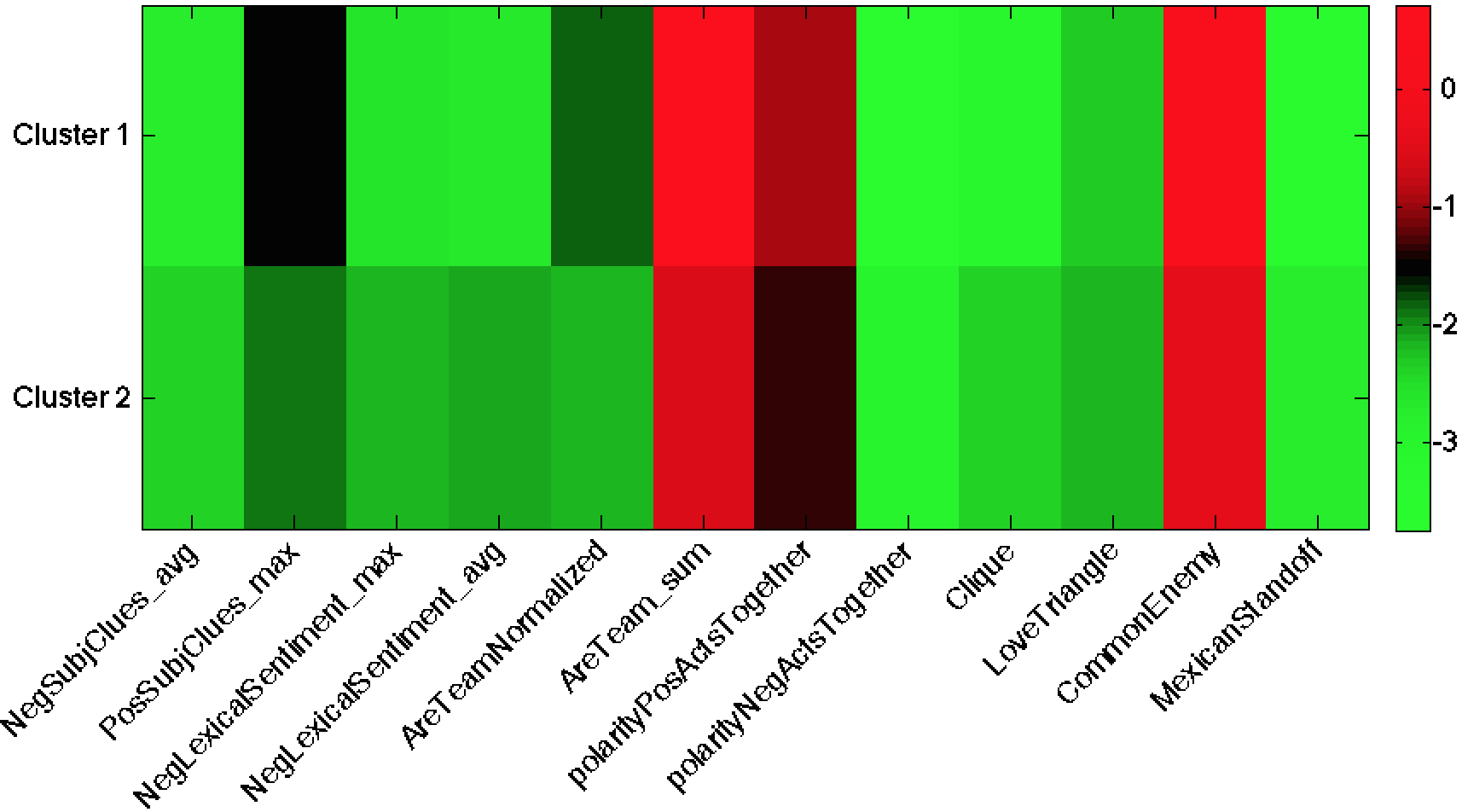}
\caption{Heatmap of cluster-weights}
\label{heatmap}
\end{figure}

\noindent \textbf{Cluster analysis:}
We briefly analyze a particular run of the clustering model for $K=2$. In Figure \ref{heatmap}, we plot the overall feature weights ($\bm{w}_k$) for a run (we plot 8 most significant features from the text model, and the primary structural features). We note that the two clusters are reasonably well delineated; and thus clustering is meaningful. For this run, Cluster 1 appears correlated with higher weights of positive polarity features. Cluster 2 appears less differentiated in terms of structural features than Cluster 1 or the non-clustering structured model.  %; and might depend primarily on text-features for predictions.

\setlength{\textfloatsep}{12pt}

\begin{figure}[h]
\centering
\includegraphics[scale =0.25]{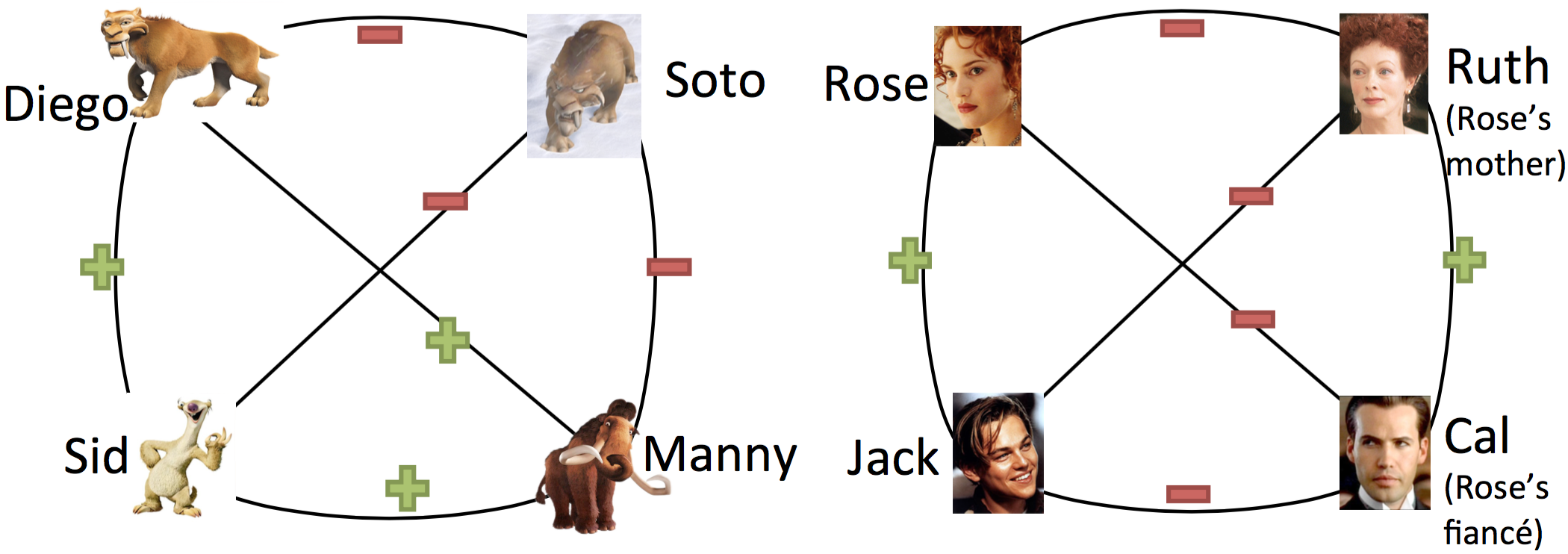}
\caption{Inferred relationships for movies `Titanic' (1997) and `Ice Age' (2002)}
\label{examples1}
\end{figure}

\noindent \textbf{Qualitative evaluation}
We observe that relation characterizations for character pairs are reasonable for most narrative texts in the test set. Figure \ref{examples1} shows labels inferred by the model for two well-known movies in the test set. Further, analysis of highest contributing evidences that lead to predictions %show that for most predictions, the model provides 
indicated that the model usually provides meaningful justifications for relationship characterization in terms of narrative acts, or implied relations from structural features.

\noindent Error analysis revealed that mismatched coreference labelings are the most common source of errors for the model. Secondly, in some cases, the text-based features mistakenly identify negative sentiments due to our coarse model of sentiment. For example, consider the following segment for the movie \emph{Smokin' Aces 2}: \emph{ `Baker \underline{drags} the wounded Redstone to the "spider trap" ... used to safeguard people'}. Here, the model mistakenly predicts the relation between Redstone and Baker as adversarial because of the negative connotation of `drag', inspite of other structural evidence. % (common adversaries). %\footnote{In this case, the model is additionally mistaken because Baker also injures Redstone in course of the narrative}

\section{Conclusion}
We have presented a framework for automatically inferring interpersonal cooperation and conflict in narrative summaries. While our testbed %we demonstrate the efficacy of our approach for plot 
was movie summaries, the framework could potentially apply to other domains of texts with social narratives, such as news stories. %, such as news-articles and literary fiction. 
Our clustering framework provides a natural approach for such domain adaptation. 
%The approach could be a useful tool to analyze business and political data, and infer relationships between companies and states. 
In the future, the framework could be extended to handle nuanced relation categorizations and asymmetric relationships. 
Conceptually, a natural extension %to the current task 
would be to use predictions about character relations to infer subtle character attributes such as agenda, intentions and goals. 
%\section*{Acknowledgments}

%\fontsize{9.5pt}{10.5pt} \selectfont
\bibliographystyle{abbrv}
\bibliography{InferringRelations}

\end{document}